\documentclass[fleqn,10pt]{wlscirep}
\usepackage[utf8]{inputenc}
\usepackage[T1]{fontenc}
\usepackage{siunitx}
\usepackage{algorithm}
\usepackage{algorithmic}
\usepackage{multirow}
\usepackage{graphicx}
\usepackage{dirtytalk}
\usepackage{subfig}
\usepackage[export]{adjustbox}

\title{Automatic Mass Detection in Breast Using Deep Convolutional Neural Network and SVM Classifier}

\author[1,*]{Md. Kamrul Hasan}
\author[1]{Tajwar Abrar Aleef}
\affil[1]{Erasmus Joint Master in Medical Imaging \& Applications (MAIA), University of Girona, Girona, Spain}

\affil[*]{md-kamrul\_hasan@etu.u-bourgogne.fr}

\keywords{Deep Convolutional Neural Networks (DCNN), SVM, Breast Cancer, Mass Classification}

\begin{abstract}
Mammography is the most widely used gold standard for screening breast cancer, where, mass detection is considered as the prominent step. Detecting mass in the breast is however an arduous problem as they usually have large variations between them in terms of shape, size, boundary, and texture. In this literature, the process of mass detection is automated with the use of transfer learning techniques of Deep Convolutional Neural Networks (DCNN). Pre-trained VGG19 network is used to extract features which are then followed by bagged decision tree for features selection and then a Support Vector Machine (SVM) classifier is trained and used for classifying between the mass and non-mass. Area Under ROC Curve (AUC) is chosen as the performance metric, which is then maximized during classifier selection and hyper-parameter tuning. The robustness of the two selected type of classifiers, C-SVM and $\upsilon$-SVM, are investigated with extensive experiments before selecting the best performing classifier. All experiments in this paper were conducted using the INbreast dataset. The best AUC obtained from the experimental results is 0.994 +/- 0.003 i.e. [0.991, 0.997]. Our results conclude that by using pre-trained VGG19 network, high-level distinctive features can be extracted from Mammograms which when used with the proposed SVM classifier is able to robustly distinguish between the mass and non-mass present in breast.

\end{abstract}
\begin{document}

\flushbottom
\maketitle

\thispagestyle{empty}

\section*{Introduction}

\begin{figure}[H]
    \centering
    \includegraphics[scale=0.65]{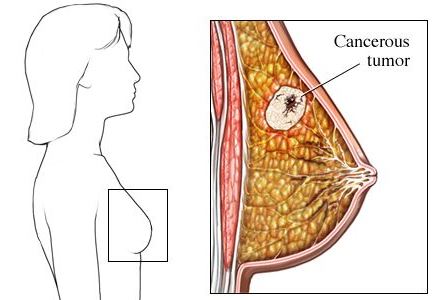}
    \caption{Presence of malignant tumor in the breast \cite{third}.}
    \label{fig:cancer}
\end{figure}

Cancer is the foremost worldwide public health problem and it is considered to be the second leading cause of death with an estimated 9.6 million deaths in 2018 \cite{cancer}. Approximately, 70 \% of cancer-related death occurs in low and middle-income countries. Cancer is a generic term for a large group of diseases that can affect any organ and can be defined as rapid growth of abnormal cells \cite{zero}. Breast cancer contributes to the second most cause of death arising due to cancer for women \cite{first}. A tumor in the breast can be defined as an uncontrolled growth of cells which can generally be of two types e.g. non-cancerous or ‘benign’, and cancerous or ‘malignant’. The term “breast cancer” refers to the presence of malignant tumor in the breast as shown in Fig. \ref{fig:cancer}. 

There are several imaging techniques available that help clinicians to analyze and pinpoint suspicious regions. Modalities such as X-ray (Mammography, Digital breast tomosynthesis, Xeromammography, Galactography), MRI, CT, PET, Ultrasound, and Scintimammography are some of the non-invasive techniques used to detect mass in the breast. Among all these methods, mammography is more commonly used and it is considered as the gold standard for detecting early stage breast cancer before the lesions become clinically palpable \cite{fourth}. In mammography, X-rays are used to produce images of the breast that provide information about breast morphology, normal anatomy, and gross pathology. Mammography is used primarily to detect and diagnose breast cancer and to evaluate palpable and non-palpable breast lesions. After mammographic imaging, one of the foremost challenges for a radiologist or a CAD system is to accurately distinguish between the mass and non-mass regions of the breast \cite{fifth}.

There are plenty of methods used in CAD systems in order to make this differentiation. In ref\cite{six}, K-means clustering, Template-matching, Simpson's Diversity Index, and finally SVM are used which resulted in an overall accuracy of the pipeline at 83.94 \%. In ref\cite{seven}, for each mass, eight shape parameters and ten enhancement texture features were calculated and then an Artificial Neural Network (ANN) was used to build the diagnostic model. The average AUC reported by the system was 0.76. In ref\cite{eight}, tissue identification results were obtained by multivariate statistical analysis of mass spectrometer data with a sensitivity of 90.9\% and specificity of 98.8\%. In ref\cite{nine}, transfer learning using AlexNet with 1656 mammography images ($454\times454$) was implemented where the system had an accuracy of 85.35 \%. Ref\cite{ten},\cite{eleven},\cite{twelve},\cite{thirteen} initially detected mass candidates or regions of interest (ROI) followed by feature vector extraction based on special knowledge and then classification was done using the feature vectors. In ref\cite{fourteen}, authors used SVMs with ConvNets to detect mass on mammograms, where the reported accuracy was 98.44\%, which is superior to the baseline (ConvNets) by 8 \%. In ref\cite{bcancer}, authors used DCNN along with SVM and were able to achieve an AUC of 0.94. In ref\cite{dm}, authors proposed a computational methodology, where, pre-processing was done initially to improve the quality of mammogram images as well as to remove regions outside the breast and hence reducing noise and highlighting internal structures of the breast. Next, cellular neural networks were used to segment the regions and to extract shape descriptors (eccentricity, circularity, density, circular disproportion, and circular density), followed by a SVM classifier. They reported a sensitivity of 80\% and AUC of 0.87. In ref\cite{dl}, pre-trained ResNet-50 architecture and Class Activation Map (CAM) technique were employed in breast cancer classification and localization resulting in an AUC of 0.96.

It is challenging to have a reliable comparison between the published methods as they are not using the same dataset. However, a general trend observed in the literature includes less focus on the computational complexity, robustness and also reporting low accuracy on unseen test set. The novelty of this research lies in the: pipeline that uses transfer learning for feature extraction, optimization process of hyper-parameters that gives the best AUC using SVM (C-SVM or $\upsilon$-SVM) classifiers, and its ability to run seamlessly on CPUs. The optimal penalization parameters C and $\upsilon$ for C-SVM and $\upsilon$-SVM were found using extensive grid search methods that maximize AUC. The proposed method in this paper uses a pre-trained VGG19 model to extract features from patches ($454\times454$) of mammogram. The features are selected using bagged decision tree to avoid redundant features in order to reduce over-fitting and the curse of dimensionality. Selected features are then given as an input to the optimized SVM classifier (C-SVM and $\upsilon$-SVM) for classification. Robustness of the classifier has been validated using 5-fold cross-validation. 

The remaining sections of this paper are organized as follows: section II is dedicated to discussing data set. Section III discusses the overall pipeline followed in this research. Section IV is about obtained results and discussions using the proposed pipeline and finally, Section V concludes and summarizes the literature.

\section*{Mammography Database}

\begin{figure}[h]

\centering
\subfloat[MLO view]{\includegraphics[width=6cm,height=6cm]{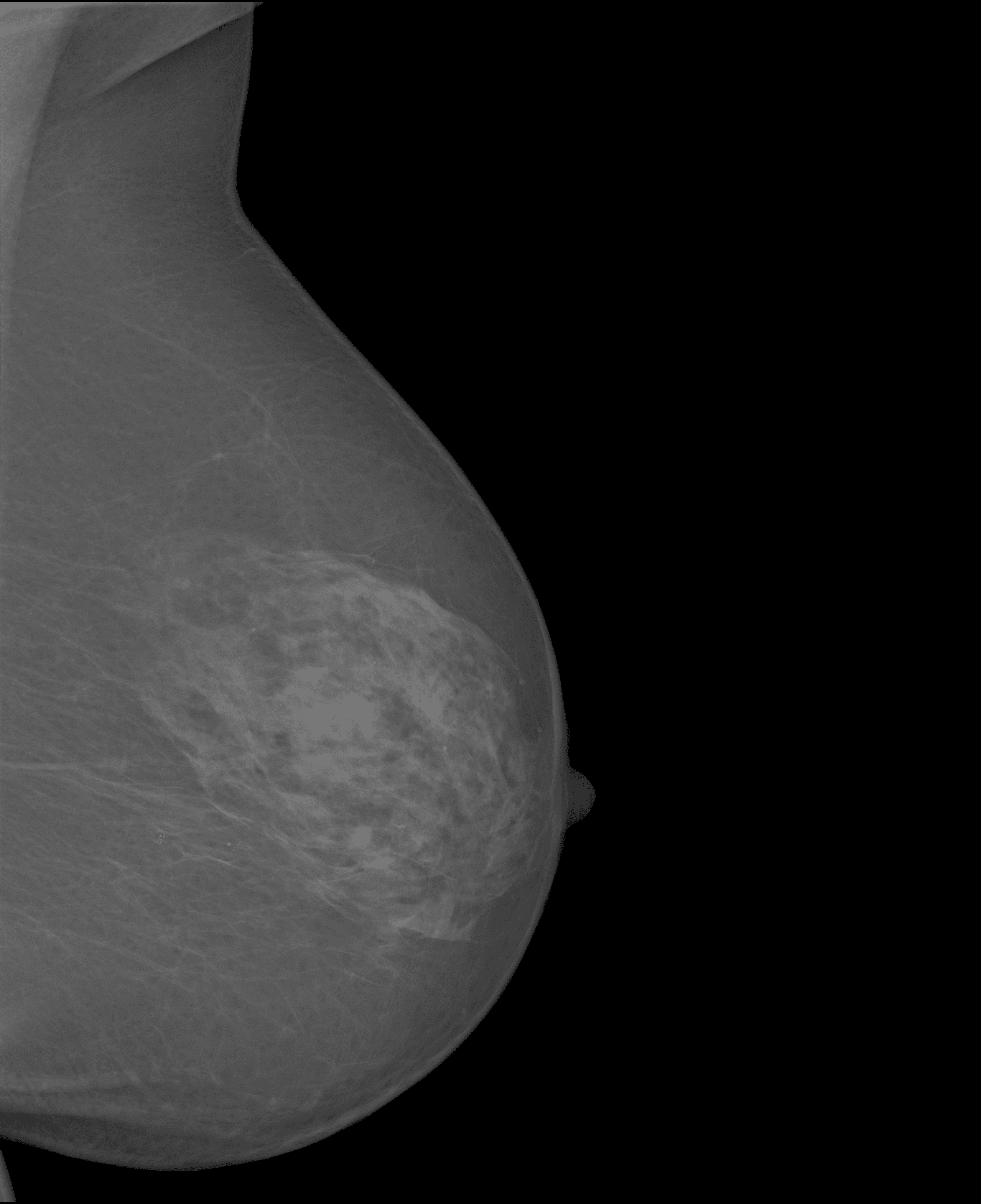}} \hfil
\subfloat[CC view]{\includegraphics[width=6cm,height=6cm]{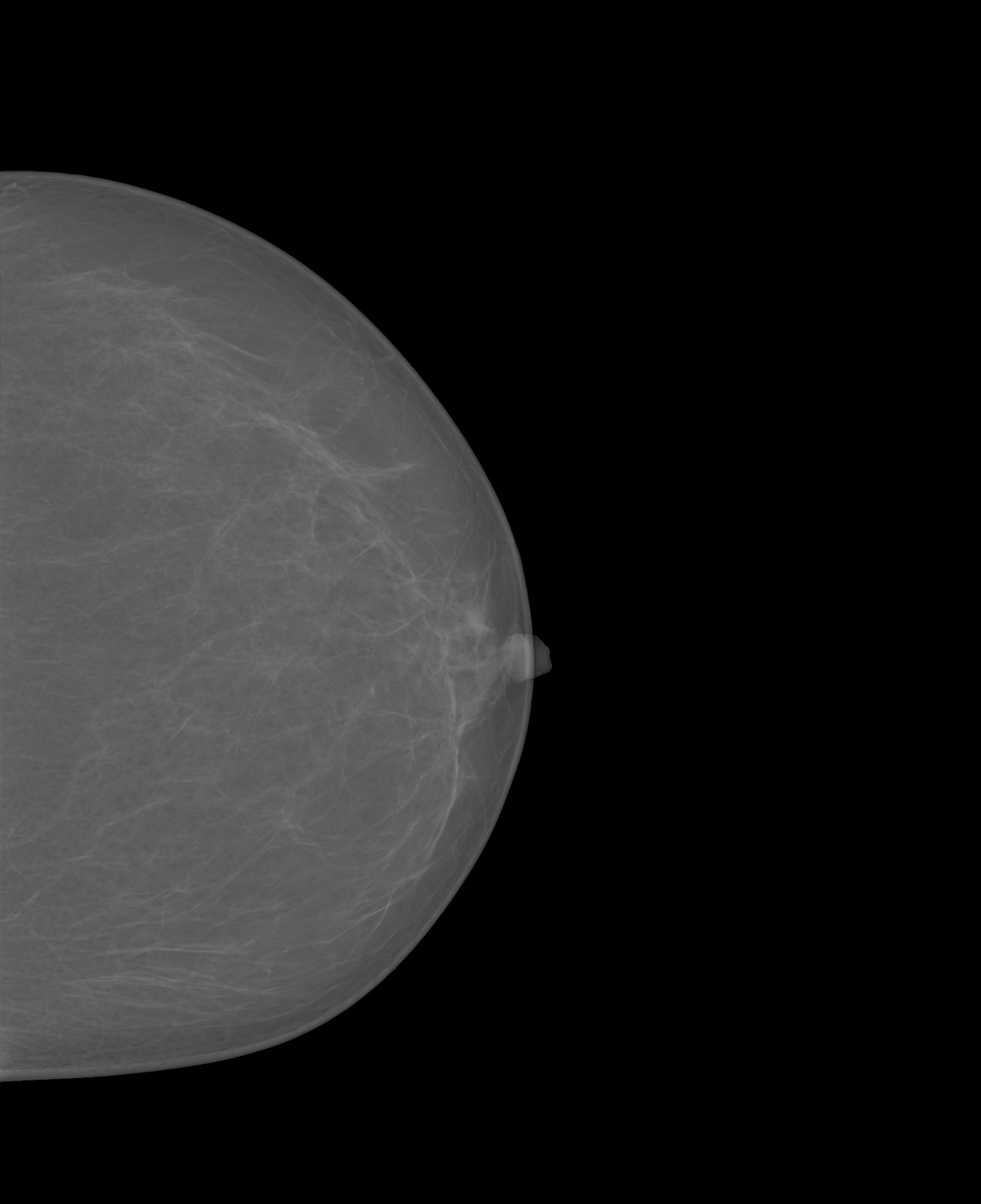}}

\caption{Two major views of Mammograms available in INbreast database.}
\label{fig:DataBase}
\end{figure}

INbreast \cite{inbreast} database was used in this literature which was acquired at a Breast Centre located in a University Hospital (Hospital de São João, Breast Centre, Porto, Portugal) and was made public for research use under the permission of both the Hospital’s Ethics Committee and the National Committee of Data Protection. INbreast mammographic images were collected using a MammoNovation Siemens FFDM with a solid-state detector of amorphous selenium. The dataset contains image dimensions of $3328\times4084$ pixels or $2560\times3328$ pixels with a pixel size of 70 microns along with a 14-bit contrast resolution. There are total of 410 images and 115 cases of which 90 of them have mediolateral oblique (MLO) view and the others have craniocaudal (CC) view. Example of the two main views, MLO (visible pectoral muscle) and CC, are shown in Fig. \ref{fig:DataBase}.

\section*{Methods}
\subsection*{Pipeline}
\begin{figure}[H]https://www.overleaf.com/project/5c7440defb0def2fb94c57a2
    \centering
    \includegraphics[scale=0.65]{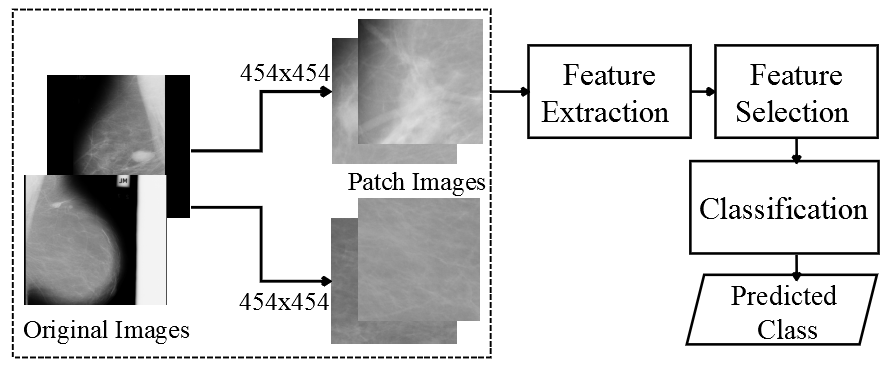}
    \caption{Overall pipeline for the proposed breast mass-detection system.}https://www.overleaf.com/project/5c7440defb0def2fb94c57a2
    \label{fig:block}
\end{figure}

The overall pipeline used for this research is shown in a block diagram as shown in Fig. \ref{fig:block}. The patch extraction step is followed by feature extraction using VGG19 network. And then, feature selection step is followed by the SVM classifier which finally classifies between the mass and non-mass tissues. Each block of the proposed pipeline is elaborately described below step by step. 

\subsection*{Patch Extraction}
Patch-based classification increases the robustness of the classifier by increasing the number of training samples \cite{patch} and reducing the computational complexity. In this study, patches having dimensions of $454\times454$ pixels are extracted from the original mammograms. To create the patches, a stride of 300 pixels was used which lowers the probability of redundant information between two consecutive patches. A total of 1000 positive patches and 1000 negative patches were extracted for training, validation, and testing. Two geometric data augmentations techniques, Flipping and Rotation, were employed to avoid over-fitting as well as the curse of dimensionality. OpenCV's python API was used for flipping the patches in x-axis and also for performing 2D rotations around row/2 and col/2 axis by \ang{90}. The following pseudocode in Algorithm \ref{alg:code} shows how the patches were extracted for training. 

\begin{algorithm}
\caption{Pseudocode for Patch Extraction}
\begin{algorithmic} 
\FOR{i=1:1:rows} 
\STATE $Final\_row=initial\_row + patch\_height$
\IF{$Final\_row < rows$}
\FOR{j=1:1:columns }
\STATE $Final\_column=initial\_column + patch\_width$
\IF {$Final\_column < columns$}
\STATE $patch=Croped(Original\ Image)$
\STATE $Saved$
\ENDIF
\STATE $initial\_column = initial\_column + stride$
\ENDFOR
\STATE $initial\_column =1$
\ENDIF
\STATE $initial\_row = initial\_row + stride$
\ENDFOR
\end{algorithmic}
\label{alg:code}
\end{algorithm}

\subsection*{Feature Extraction using VGG19}
\begin{figure}[H]
    \centering
    \includegraphics[width=17.5cm,height=12cm,keepaspectratio]{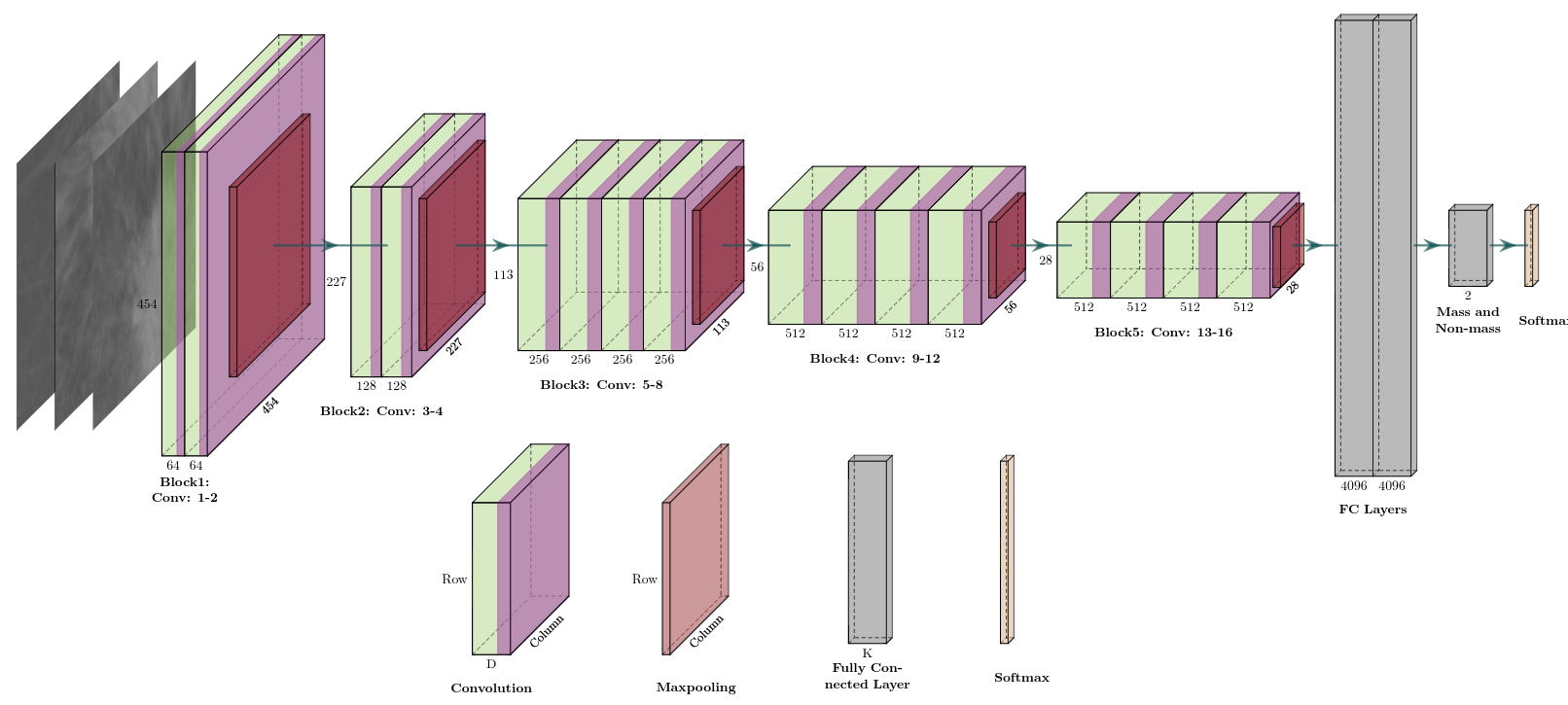}
    \caption{Details of the 19 layers of VGG19 network \cite{vgg19} used for feature extraction.}
    \label{fig:vgg19}
\end{figure}

After extracting the patches, feature extraction was conducted by passing the patches through the pre-trained VGG19 network\cite{vgg19}. The pre-trained VGG19  model has a depth of 19 layers and was trained using the ImageNet dataset\cite{imagenet}. VGG19 network can be characterized by its simplicity that uses only $3\times3$ convolutional layers stacked on top of each other in increasing depths. To reduce dimensionality, down-sampling of input representation (image, hidden-layer output matrix, etc) is used in this network. Two fully-connected layers (FC1 and FC2), each with 4096 nodes are then followed by a softmax layer which in together forms the classifier. The adopted VGG19 \cite{vgg19} network with 19 layers is shown in detail in Fig. \ref{fig:vgg19}.



Patches were normalized with zero mean and unit variance before feeding them into the network for feature extraction. Two sets of features were extracted from the VGG19 model, first one after the fully connected layer 2 (FC2) that gives 4096 features and second one after the flatten layer (FL) that gives 25088 features. These two sets of features were later used to compare the effect of the dimension of features vectors on the robustness and computational complexity of final classifier. 


\subsection*{Feature Selection}
\begin{figure}[H]
    \centering
    \includegraphics[scale=0.75]{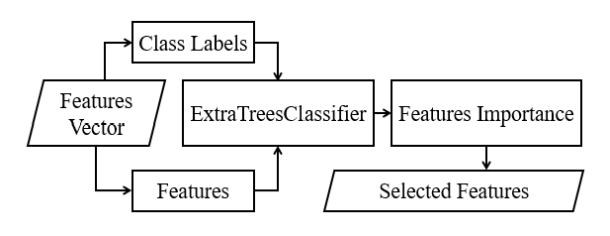}
    \caption{Flow diagram of the implemented feature selector.}
    \label{fig:f_selection}
\end{figure}

Not all extracted features are relevant, reducing redundant and irrelevant features benefits the classifier in a couple of ways. It i) reduces over-fitting, ii) improves accuracy, iii) reduces training time, and iv) decreases the complexity of the classifier. To select the optimum set of features, bagged decision trees like Random Forest and Extra Trees were used. The bagged algorithm \cite{bagged} implements a meta estimator that fits several randomized decision trees (also known as extra-trees) on various sub-samples of the dataset. In this literature, Random Forest and Extra Trees are used to estimate the importance of feature and then the features were selected depending on their rank. To implement features selector, \say{ExtraTreesClassifier} class in the \say{scikit-learn} API was used. To select the features from the feature importance, 95\% importance was considered as mentioned in En. \ref{eq:1}. The working flow diagram of the implemented feature selection technique is shown in Fig. \ref{fig:f_selection}. 

\begin{equation} \label{eq:1} \centering 
Selected\;Feature\;importance = 0.95 \times Total\; importance
\end{equation}


\subsection*{Train, Validation and Test Data Selection}
\begin{figure}[H]
    \centering
    \includegraphics[scale=0.55]{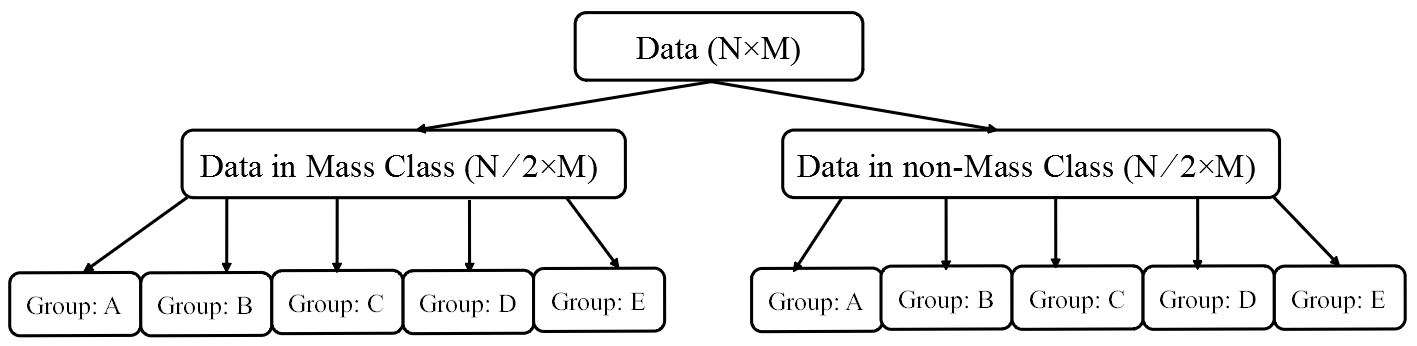}
    \caption{Mass and Non-mass data split for the cross validation.}
    \label{fig:datasplit}
\end{figure}

After selecting the optimum set of features, this feature vector was split into train, validation and test data. To do so, firstly, whole data was divided into five (5-fold cross validation \cite{kcv}) equal parts as shown in Fig. \ref{fig:datasplit}. Cross-fold validation was done in order to get a more generalized model that can provide results which can be expected from unseen test data. $N$ is the total number of observations in both classes (mass and non-mass) and $M$ is the number of selected features where data was balanced in both the mass and non-mass classes. Data having an equal number of classes was chosen randomly: 60\% for training, 20 \% for validation and 20 \% for testing.


From Fig. \ref{fig:datasplit}, it is seen that there are 5 possible cases of train, validation and test data split as given in Table \ref{tab:datatable}. For each case, AUC for test data was calculated after tuning the classifier's hyper-parameters from the validation stage. 

\begin{table}[h]
\centering
\caption{Training, validation and test selection}
\label{tab:datatable}
\begin{tabular}{|c|l|l|l|l|l|l|}
\hline
\multirow{2}{*}{Possible Cases} & \multicolumn{2}{c|}{Train Set (60\%)}  & \multicolumn{2}{c|}{Validation Set (20\%)} & \multicolumn{2}{c|}{Test Set (20\%)}  \\ \cline{2-7} 
 & Mass class  & Non-mass class & Mass class  & Non-mass class & Mass class  & Non-mass class\\ \hline
Case I & Group: A, B, C & Group: A, B, C & Group: D & Group: D & Group: E & Group: E\\ \hline
Case II & Group: B, C, D & Group: B, C, D & Group: E & Group: E & Group: A & Group: A\\ \hline
Case III & Group: C, D, E & Group: C, D, E & Group: A & Group: A & Group: B & Group: B\\ \hline
Case IV & Group: D, E, A & Group: D, E, A & Group: B & Group: B & Group: C & Group: C\\ \hline
Case V & Group: E, A, B & Group: E, A, B & Group: C & Group: C & Group: D & Group: D\\ \hline
\end{tabular}
\end{table}

\subsection*{Performance Metric for SVM}
In statistics, a Receiver Operating Characteristic (ROC)\cite{rocauc} curve is a graphical plot that illustrates the diagnostic (classification) ability of a binary classifier system as its discrimination threshold is varied which is considered as a fundamental tool for diagnostic test evaluation. In a ROC curve, the True Positive Rate (TPR) i.e. Sensitivity is plotted in function of the False Positive Rate (FPR) (100-Specificity) for different cut-off points. Each point on the ROC curve represents a sensitivity/ specificity pair. Lowering the classification threshold classifies more items as positive, thus increasing both False Positives (FP) and True Positives (TP). The AUC\cite{rocauc} is a measure of how well a parameter can distinguish between two diagnostic class which ranges between $0\sim1$. Alternately, AUC provides an aggregate measure of performance across all possible classification thresholds. In this paper, standard deviation was used to measure the robustness of the classifier, where, a small value of standard deviation indicates that classifier is more generalized and robust on independent unseen data. Robustness of the classifier means it is possible to reproduce the output class posterior probability with a minimum level of discrepancy.

\subsection*{Classification using Support Vector Machine (SVM)}
Support vector machines (SVMs), originally proposed by V. Vapnik, have been applied to many classification problems in medical imaging. However, classes having larger training samples have a bias on the decision boundary \cite{svmbiased}. This unwanted class bias can be penalized using C-SVM and $\upsilon$-SVM by introducing penalty parameter C and $\upsilon$ respectively. In C-SVM and $\upsilon$-SVM, C and $\upsilon$ both are regularization parameters which support to implement a penalty on the miss-classifications. After splitting features vector in train, validation and test set, SVM \cite{svm} (C-SVM and $\upsilon$-SVM) classifiers are trained and compared in order to select the best classifier for the prediction of mass and non-mass patches.



\begin{figure}[H]
    \centering
    \includegraphics[scale=0.59]{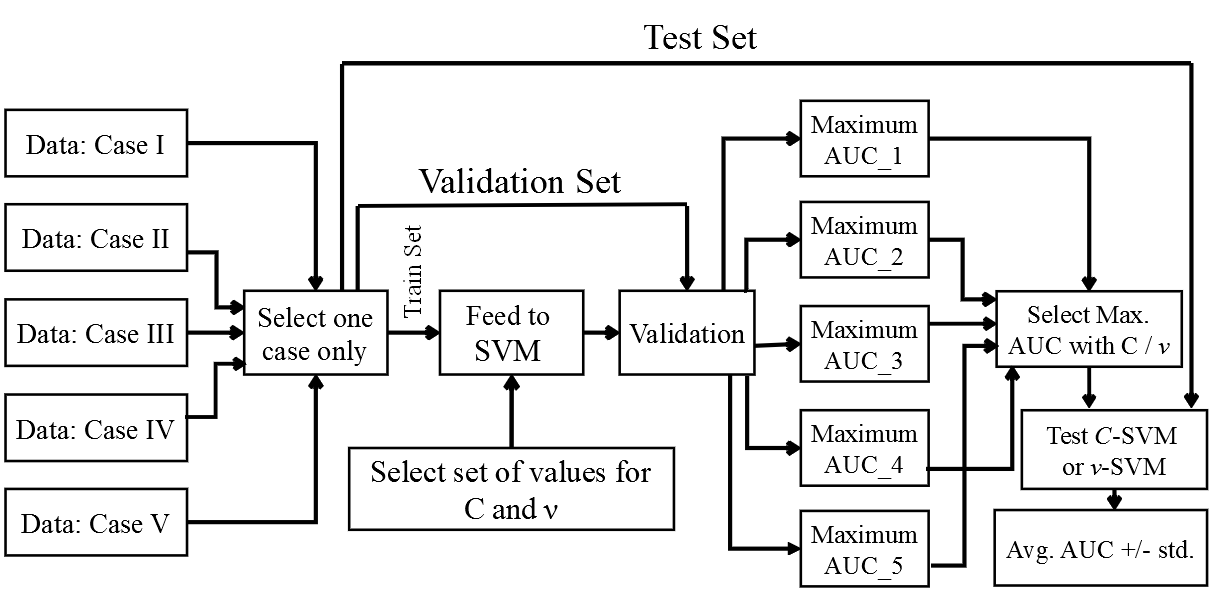}
    \caption{Block diagram for tuning the best hyper-parameters and predicting results on test set.}
    \label{fig:workflow}
\end{figure}

To select the best hyper-parameters e.g. C and $\upsilon$ for C-SVM and $\upsilon$-SVM, cross-validation was used. To do hyper-parameters tuning (optimization), grid search \cite{gridsearch} was used which can be defined simply as an exhaustive searching through a manually specified subset of the hyper-parameter space of a learning algorithm. A grid search algorithm should be steered by some performance metric which was the average AUC in our case. The overall workflow for getting best hyper-parameters and test AUC is shown in Fig. \ref{fig:workflow}. 

From Fig. \ref{fig:workflow}, it is seen that from the set of penalization parameters (C and $\upsilon$), grid search automatically return the best C and $\upsilon$ values for C-SVM and $\upsilon$-SVM that maximizes AUC based on the validation sets. After selecting the best values for C and $\upsilon$, the network was evaluated using the corresponding test data. The five different values of AUC is then obtained for the five different cases. The average AUC with standard deviation was calculated and compared from the 5 fold cross-validation sets. All experiments were performed in a Windows machine (Intel (R) Core (TN) i5-7200U CPU @ 2.5 GHz 2.71 GHz). System type was 64-bit x64-based processor with 8 GB of RAM.

%

\section*{Results \& Discussions}
In all the experiments with C-SVM and $\upsilon$-SVM, only results with radial basis function (RBF)\cite{rbf} kernel are reported in this paper as it had the best performance compared to using polynomial, linear or sigmoid kernel after calibration. The research results described in this section is for C=$10^{-3}\sim10^4$  and  $\upsilon$=$0.001\sim0.9$ and these values of C and  $\upsilon$ were optimized in the validation stage by maximizing AUC. For all these experiments, average AUC with standard deviation for test data and total computation time (features extraction, training, penalization parameters optimization, and testing of SVM model) were recorded. Some extracted patches and their corresponding augmented (flipping and rotation) images are given in Table \ref{tab:patches}. The experiments performed in this research for obtaining the best classifier are as follows: 

\begin{table}[H]
\setlength\tabcolsep{1.5pt}
\caption{Example of extracted Patches with corresponding geometric augmentation}
\centering
\begin{tabular}{c|c|c|||c|c|c|}
Positive patch & Flipped Positive & Rotated Positive & Negative Patch & Flipped Negative & Rotated Negative   \\ \hline

\includegraphics[scale=0.15,valign=c]{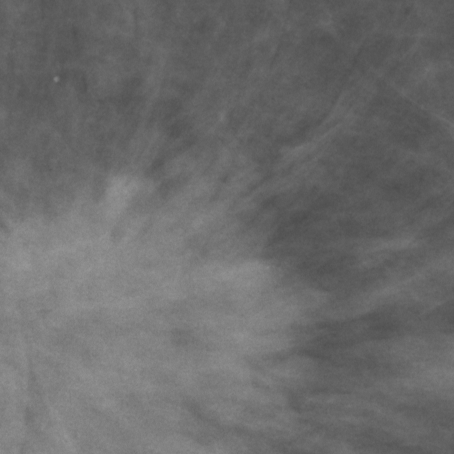}& 
\includegraphics[scale=0.15,valign=c]{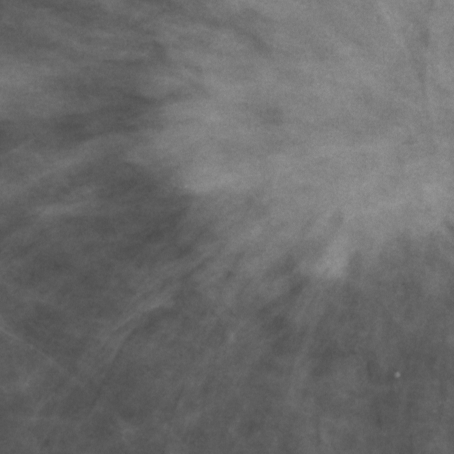}& 
\includegraphics[scale=0.15,valign=c]{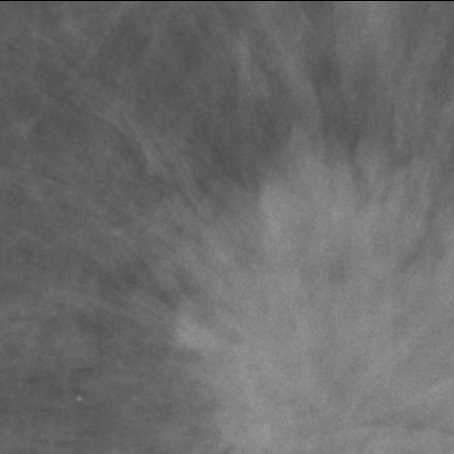}& 
\includegraphics[scale=0.15,valign=c]{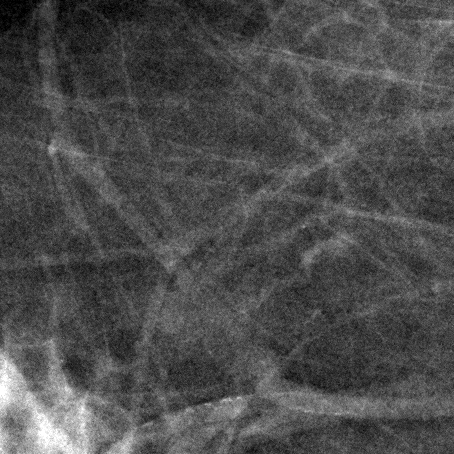}&
\includegraphics[scale=0.15,valign=c]{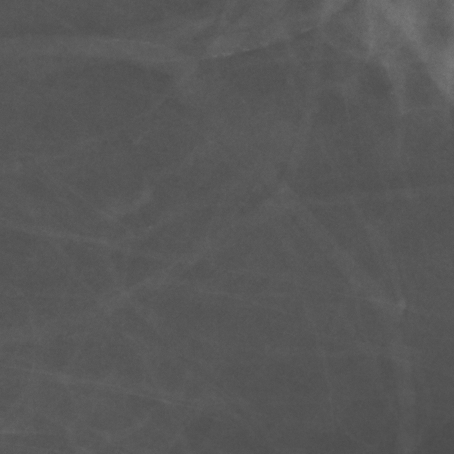}&
\includegraphics[scale=0.15,valign=c]{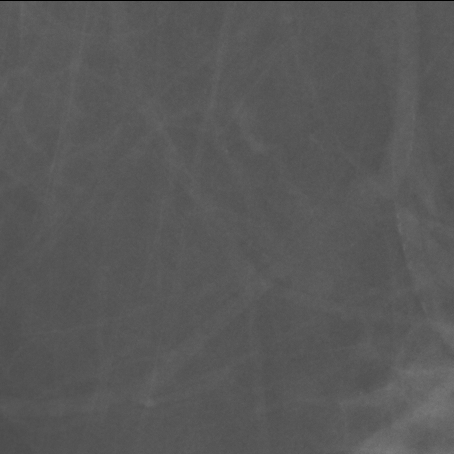}
 
\\  \hline

\includegraphics[scale=0.15,valign=c]{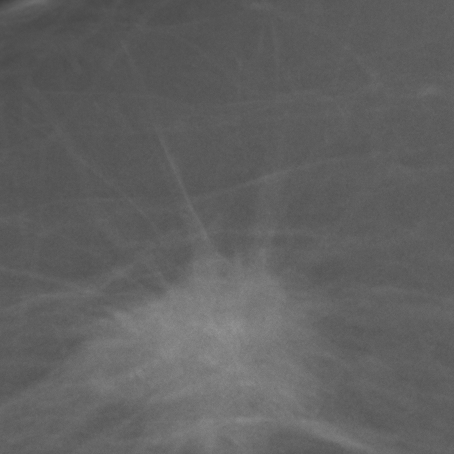}& 
\includegraphics[scale=0.15,valign=c]{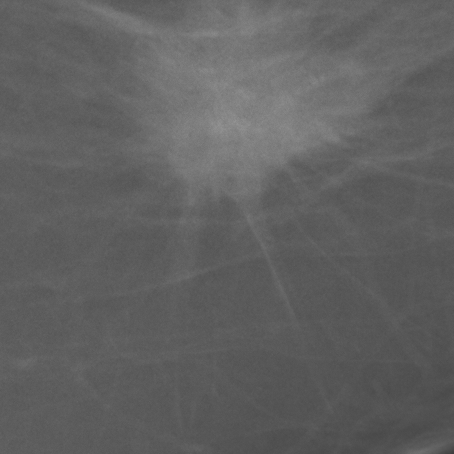}& 
\includegraphics[scale=0.15,valign=c]{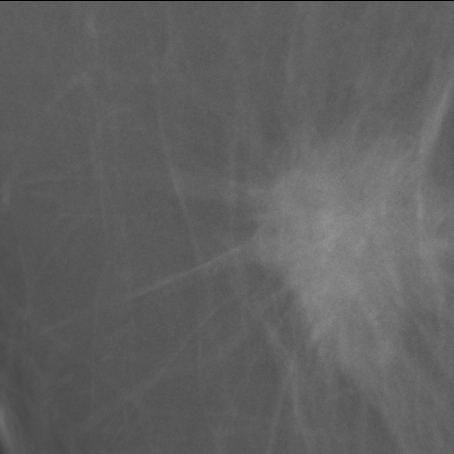}& 
\includegraphics[scale=0.15,valign=c]{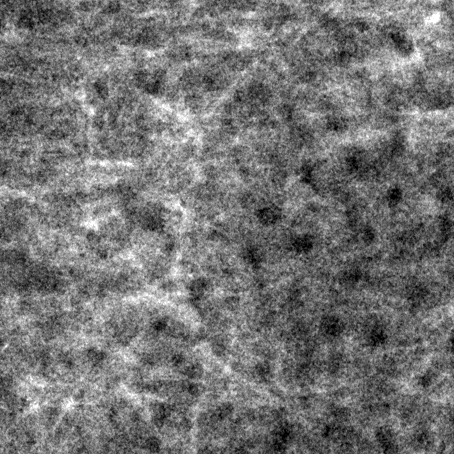}&
\includegraphics[scale=0.15,valign=c]{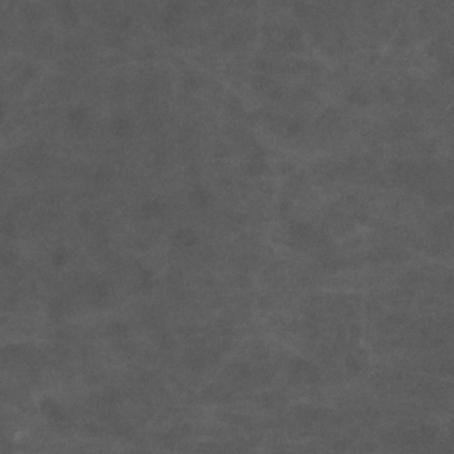}& 
\includegraphics[scale=0.15,valign=c]{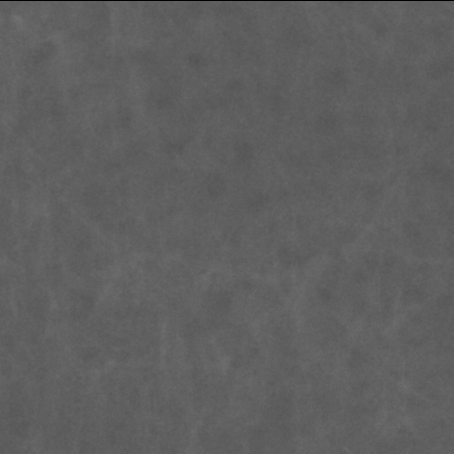}
 
\\  \hline
 
\end{tabular}
\label{tab:patches}
\end{table}

\subsection*{Experiment 1}
In this experiment, the features come from fully connected layer 2 (FC2) of VGG19 network that has 4096 features and using 2000 instances of observations for both classes. Among 4096 features, 90 features that satisfy the En. \ref{eq:1} are considered in this experiment. From validation set, it is found that maximum AUC is at C = 100 and $\upsilon$ = 0.01. Using those values, the best model was selected to examine on the unseen test data. Test results for C-SVM and $\upsilon$-SVM are given in Table \ref{tab:case_1}.

\begin{table}[H]
\centering
\caption{Test results for C-SVM (Left) and $\upsilon$-SVM (Right) from Experiment 1}
\label{tab:case_1}
\begin{tabular}{|c|c|c|c|c|}
\hline
\multicolumn{1}{|l|}{\multirow{2}{*}{\textbf{Test Data}}} & \multicolumn{2}{c|}{\textbf{C-SVM}} & \multicolumn{2}{c|}{\textbf{$\upsilon$-SVM}} \\ \cline{2-5} 
\multicolumn{1}{|l|}{} & \textbf{AUC} & \textbf{Average AUC} & \textbf{AUC} & \textbf{Average AUC} \\ \hline
\textbf{Case I} & 0.95905 & \multirow{5}{*}{0.989 (+/-0.015)} & 0.95800 & \multirow{5}{*}{0.988 (+/-0.016)} \\ \cline{1-2} \cline{4-4}
\textbf{Case II} & 0.99002 &  & 0.99025 &  \\ \cline{1-2} \cline{4-4}
\textbf{Case III} & 0.99900 &  & 0.99905 &  \\ \cline{1-2} \cline{4-4}
\textbf{Case IV} & 0.99495 &  & 0.99482 &  \\ \cline{1-2} \cline{4-4}
\textbf{Case V} & 1.00000 &  & 0.99995 &  \\ \hline
\end{tabular}
\end{table}

\begin{figure}[h]

\centering
\subfloat[For C-SVM]{\includegraphics[width=8.5cm,height=6.25cm]{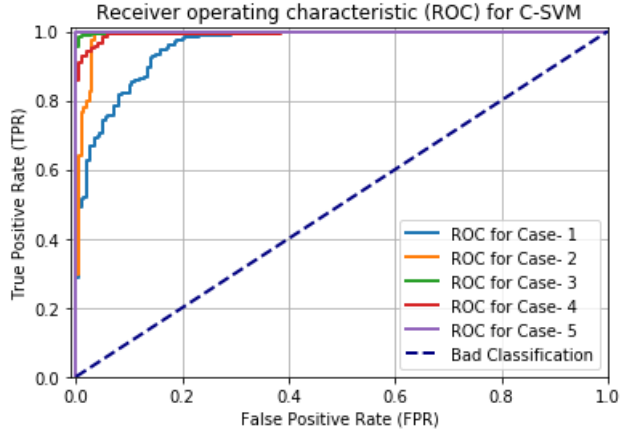}} \hfil
\subfloat[For $\upsilon$-SVM]{\includegraphics[width=8.5cm,height=6.25cm]{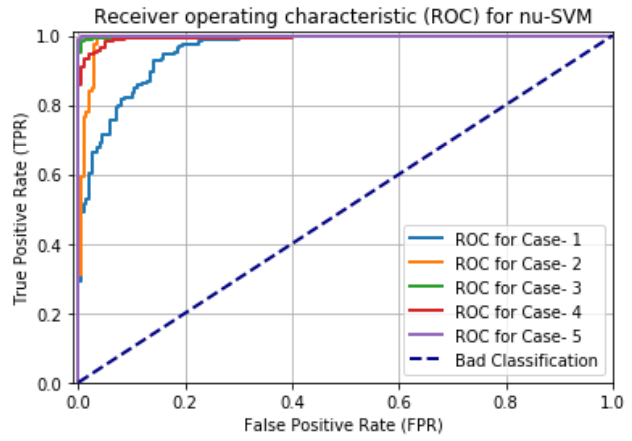}}

\caption{ROC curves for test data of Experiment 1.}
\label{fig:case1}
\end{figure}

From Table \ref{tab:case_1}, it is seen that average AUC is 0.989 +/- 0.015 i.e. [0.974, 1.00] for C-SVM and 0.988 +/- 0.016 i.e. [0.972, 1.00] for $\upsilon$-SVM respectively. The ROC curves for C-SVM and $\upsilon$-SVM are shown in Fig. \ref{fig:case1} (left) and Fig. \ref{fig:case1} (right). Total computation time which includes the feature extraction, feature selection and classification stage was 11 mins using the CPU machine. Although, both the penalization parameters, C and $\upsilon$ are obtained from the grid search optimizer using the same set of validation data, from Table \ref{tab:case_1} and Fig. \ref{fig:case1}, it is noticed that C-SVM performs better than the $\upsilon$-SVM.

\subsection*{Experiment 2}
In this experiment, the features were obtained from the flatten layer (FL) of VGG19 network that has 25088 features and using 2000 instances of the observations. Total of 189 features have been selected in this experiment. From the validation set, it is found that maximum AUC is at C = 10 and $\upsilon$ = 0.1. Best SVM model was taken using those values of C and $\upsilon$ for test data. The obtained test results for C-SVM and $\upsilon$-SVM are given in Table \ref{tab:case_2}.

\begin{table}[H]
\centering
\caption{Test results for C-SVM (Left) and $\upsilon$-SVM (Right) for Experiment 2}
\label{tab:case_2}
\begin{tabular}{|c|c|c|c|c|}
\hline
\multicolumn{1}{|l|}{\multirow{2}{*}{\textbf{Test Data}}} & \multicolumn{2}{c|}{\textbf{C-SVM}} & \multicolumn{2}{c|}{\textbf{$\upsilon$-SVM}} \\ \cline{2-5} 
\multicolumn{1}{|l|}{} & \textbf{AUC} & \textbf{Average AUC} & \textbf{AUC} & \textbf{Average AUC} \\ \hline
\textbf{Case I} & 0.916975 & \multirow{5}{*}{0.976 (+/-0.031)} & 0.91697 & \multirow{5}{*}{0.976 (+/-0.031)} \\ \cline{1-2} \cline{4-4}
\textbf{Case II} & 0.971325 &  & 0.97132 &  \\ \cline{1-2} \cline{4-4}
\textbf{Case III} & 0.999224 &  & 0.99922 &  \\ \cline{1-2} \cline{4-4}
\textbf{Case IV} & 0.995400 &  & 0.99540 &  \\ \cline{1-2} \cline{4-4}
\textbf{Case V} & 0.997675 &  & 0.99767 &  \\ \hline
\end{tabular}
\end{table}

\begin{figure}[h]

\centering
\subfloat[For C-SVM]{\includegraphics[width=8.5cm,height=6.25cm]{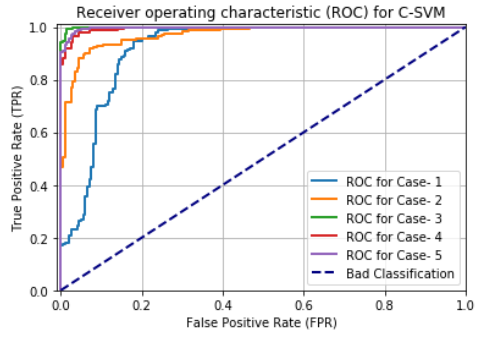}} \hfil
\subfloat[For $\upsilon$-SVM]{\includegraphics[width=8.5cm,height=6.25cm]{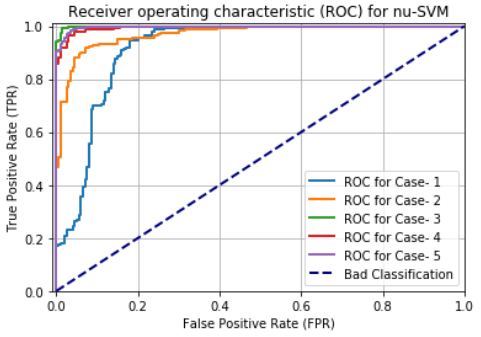}}

\caption{ROC curves for test data of Experiment 2.}
\label{fig:case2}
\end{figure}

It is seen from Table \ref{tab:case_2} that average AUC is 0.976 +/-0.031 i.e. [0.945, 1.00] and 0.976 +/-0.031 i.e. [0.945, 1.00] for C-SVM and $\upsilon$-SVM respectively. The ROC curves for C-SVM and $\upsilon$-SVM are shown in Fig. \ref{fig:case2} (left) and Fig. \ref{fig:case2} (right). Total computation time now was 40 mins. From Experiment 2, it can be seen that both the C-SVM and $\upsilon$ are giving similar results. The performance of Experiment 1 is slightly better than the performance of Experiment 2 even though both the experiments have the same number of observations. This can be explained by the fact that Experiment 2 uses a higher number of features which can result in overfitting. 

\subsection*{Experiment 3}

\begin{figure}[h]

\centering
\subfloat[For C-SVM]{\includegraphics[width=8.5cm,height=6.25cm]{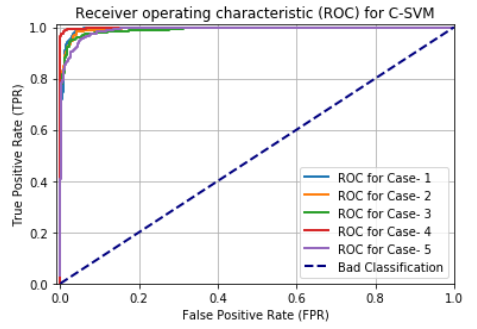}} \hfil
\subfloat[For $\upsilon$-SVM]{\includegraphics[width=8.5cm,height=6.25cm]{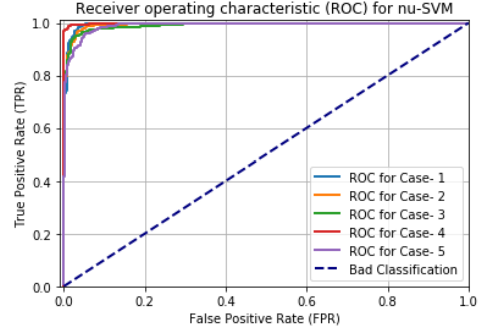}}

\caption{ROC curves for test data of Experiment 3.}
\label{fig:case3}
\end{figure}

In this experiment, two geometric augmentations (flipping and rotation) was added on extracted patches and the features were extracted from fully connected layer 2 (FC2) of VGG19 network which gave 4096 features and now with augmentation, there were 6000 instances of the observations for this experiment. The selected feature number is 97 in this experiment. From the validation test, it is seen that maximum AUC is at C= 100 and $\upsilon$=0.005. Using these values, the best SVM model was selected to examine the test data. Test results for C-SVM and $\upsilon$-SVM are given in Table \ref{tab:case_3}. Average AUC is 0.994 +/-0.003 i.e. [0.991, 0.997] and 0.994 +/-0.003 i.e. [0.991, 0.997] for C-SVM and $\upsilon$-SVM respectively. The ROC curves for C-SVM and $\upsilon$-SVM are shown in Fig. \ref{fig:case3} (left) and Fig. \ref{fig:case3} (right). Total computation time for this experiment was 3 hrs18 mins. Although the computation time here is more than the first two experiments, the performance is much better than the previous two experiments. From Fig. \ref{fig:case3} and Table \ref{tab:case_3}, it can clearly be seen that for both C-SVM and $\upsilon$-SVM and for all the test cases, the ROC curves and AUC's are quite similar. This shows the robustness of this method compared to the first two experiments. 

\begin{table}[H]
\centering
\caption{Test results for C-SVM (Left) and $\upsilon$-SVM (Right) for Experiment 3}
\label{tab:case_3}
\begin{tabular}{|c|c|c|c|c|}
\hline
\multicolumn{1}{|l|}{\multirow{2}{*}{\textbf{Test Data}}} & \multicolumn{2}{c|}{\textbf{C-SVM}} & \multicolumn{2}{c|}{\textbf{$\upsilon$-SVM}} \\ \cline{2-5} 
\multicolumn{1}{|l|}{} & \textbf{AUC} & \textbf{Average AUC} & \textbf{AUC} & \textbf{Average AUC} \\ \hline
\textbf{Case I} & 0.99521 & \multirow{5}{*}{0.994 (+/-0.003)} & 0.99537 & \multirow{5}{*}{0.994 (+/-0.003))} \\ \cline{1-2} \cline{4-4}
\textbf{Case II} & 0.99453 &  & 0.99441 &  \\ \cline{1-2} \cline{4-4}
\textbf{Case III} & 0.99079 &  & 0.99129 &  \\ \cline{1-2} \cline{4-4}
\textbf{Case IV} & 0.99893 &  & 0.99906 &  \\ \cline{1-2} \cline{4-4}
\textbf{Case V} & 0.99109 &  & 0.99129 &  \\ \hline
\end{tabular}
\end{table}

\subsection*{Experiment 4}
In this experiment, again the same two geometric augmentations (flipping and rotation) were executed on extracted patches and the features were extracted from the flatten layer (FL) of the network which provides 25088 features. In this experiment, a total of 205 features were selected. From the validation test, it is seen that maximum AUC is at C=10 and $\upsilon$=0.001. Using these values of the hyper-parameters, best model was selected to validate the test data. Test results for C-SVM and $\upsilon$-SVM are given in Table \ref{tab:case_4}. Now, the average AUC is 0.988 +/- 0.009 i.e. [0.991, 0.997] and 0.988 +/- 0.009 i.e. [0.991, 0.997] for C-SVM and $\upsilon$-SVM respectively.

\begin{table}[H]
\centering
\caption{Test results for C-SVM (Left) and $\upsilon$-SVM (Right) for Experiment 4}
\label{tab:case_4}
\begin{tabular}{|c|c|c|c|c|}
\hline
\multicolumn{1}{|l|}{\multirow{2}{*}{\textbf{Test Data}}} & \multicolumn{2}{c|}{\textbf{C-SVM}} & \multicolumn{2}{c|}{\textbf{$\upsilon$-SVM}} \\ \cline{2-5} 
\multicolumn{1}{|l|}{} & \textbf{AUC} & \textbf{Average AUC} & \textbf{AUC} & \textbf{Average AUC} \\ \hline
\textbf{Case I} & 0.98625 & \multirow{5}{*}{0.988 (+/-0.009)} & 0.98636 & \multirow{5}{*}{0.988 (+/-0.009)} \\ \cline{1-2} \cline{4-4}
\textbf{Case II} & 0.97621 &  & 0.97550 &  \\ \cline{1-2} \cline{4-4}
\textbf{Case III} & 0.98049 &  & 0.98117 &  \\ \cline{1-2} \cline{4-4}
\textbf{Case IV} & 0.99823 &  & 0.99820 &  \\ \cline{1-2} \cline{4-4}
\textbf{Case V} & 0.99704 &  & 0.99688 &  \\ \hline
\end{tabular}
\end{table}

\begin{figure}[h]

\centering
\subfloat[For C-SVM]{\includegraphics[width=8.5cm,height=6.25cm]{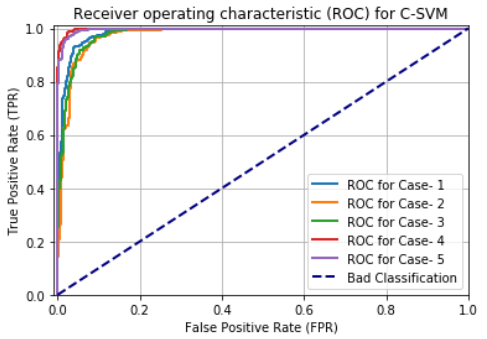}} \hfil
\subfloat[For $\upsilon$-SVM]{\includegraphics[width=8.5cm,height=6.25cm]{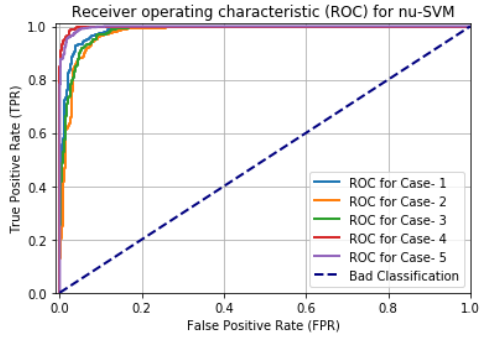}}

\caption{ROC curves for test data of Experiment 4.}
\label{fig:case4}
\end{figure}

The ROC curves for C-SVM and $\upsilon$-SVM are shown in Fig. \ref{fig:case4} (left) and Fig. \ref{fig:case4} (right). Total computation time for this experiment was 12 hrs. From Fig. \ref{fig:case4}, it is observed that the ROC behavior of all the cases of the test data in this approach is better than the first two experiments with a significant increase in the computation time. 

\section*{Conclusions}



In this literature, the robustness of the VGG19 along with SVM (C-SVM and $\upsilon$-SVM) for the breast mass and non-mass classification was analyzed. All the parameters and hyperparameters of the C-SVM and $\upsilon$-SVM (except the penalization parameters), the number of observations, and the dimensions of the features vector were kept the same during each of the experiment. For each experiment, the test results were obtained using the best C value for C-SVM and best $\upsilon$ value for $\upsilon$-SVM using extensive grid search algorithm. From the experimental results, it is worth mentioning that Experiment 4 gives higher results than the first two experiments. The observed reason can be pointed as Experiment 4 used a higher number of the observations which is three times more than the first two experiments. But, computationally, 12 hrs is excessively higher compared to all the other experiments. On the other-hand, Experiment 3 has the best results than all the other experiments with an AUC of 0.994 +/-0.003 i.e. [0.991, 0.997] and computation time of 3 hrs 18 mins. From the ROC curves of all the experiments, qualitatively it can be seen that Experiment 3 is more robust than others due to having almost similar ROC for all the test cases (for each fold). From the experimental tables, quantitatively, it is seen that Experiment 3 has less value of standard deviation which also quantitatively proves that this method is more robust than others. Same performance (AUC) for both C-SVM and $\upsilon$-SVM are achieved in experiment 3 using grid search algorithm for the hyper-parameters optimization which resulted in values for C and $\upsilon$ at100 and 0.005. Either one of the C-SVM and $\upsilon$-SVM method can be selected for the future applications for mass and non-mass breast tissue classification as they are expected to give similar results using the trained classifier from experiment 3. We conclude that even with a small training set, it is possible to obtain a robust classifier for the mass and non-mass tissue classification in the breast using our proposed pipeline. \\ 

\bibliography{sample}





\end{document}